\documentclass[10pt,twocolumn,letterpaper]{article}

\usepackage{cvpr}              
\usepackage{multirow}
\usepackage{comment}
\usepackage{siunitx}

\usepackage{pifont}

\definecolor{cvprblue}{rgb}{0.21,0.49,0.74}
\usepackage[pagebackref,breaklinks,colorlinks,allcolors=cvprblue]{hyperref}

\def\paperID{9053} 
\def\confName{CVPR}
\def\confYear{2026}

\title{Enriching Knowledge Distillation with Cross-Modal Teacher Fusion}

\author{
Amir M. Mansourian \quad Amir Mohammad Babaei \quad Shohreh Kasaei\\
Image Processing Lab, Sharif University of Technology\\
Tehran, Iran\\
{\tt\small \{amir.mansurian, amir.babaei79, kasaei\}@sharif.edu}
}

\begin{document}
\maketitle
\thispagestyle{plain}
\pagestyle{plain}   

\begin{abstract}
Multi-teacher knowledge distillation (KD), a more effective technique than traditional single-teacher methods, transfers knowledge from expert teachers to a compact student model using logit or feature matching. However, most existing approaches lack knowledge diversity, as they rely solely on unimodal visual information, overlooking the potential of cross-modal representations. In this work, we explore the use of CLIP’s vision–language knowledge as a complementary source of supervision for KD, an area that remains largely underexplored. We propose a simple yet effective framework that fuses the logits and features of a conventional teacher with those from CLIP. By incorporating CLIP’s multi-prompt textual guidance, the fused supervision captures both dataset-specific and semantically enriched visual cues. Beyond accuracy, analysis shows that the fused teacher yields more confident and reliable predictions, significantly increasing confident-correct cases while reducing confidently wrong ones. Moreover, fusion with CLIP refines the entire logit distribution, producing semantically meaningful probabilities for non-target classes, thereby improving inter-class consistency and distillation quality. Despite its simplicity, the proposed method, En\textbf{Rich}ing \textbf{K}nowledge \textbf{D}istillation (RichKD), consistently outperforms most of existing baselines across multiple benchmarks and exhibits stronger robustness under distribution shifts and input corruptions. 
\end{abstract}    
\section{Introduction}

Knowledge Distillation (KD)~\cite{gou2021knowledge, mansourian2025comprehensive} has become one of the most effective strategies for compressing deep neural networks, enabling lightweight student models to inherit the representational power of larger teachers. By transferring knowledge from a well-trained teacher through logit or feature alignment, KD has achieved success across image recognition, object detection, and many other domains. Despite these advances, most existing KD methods remain strictly \emph{unimodal}, relying solely on visual cues learned from the target dataset. As a result, the distilled knowledge often inherits the teacher’s task-specific biases and lacks the semantic diversity necessary for broader generalization.

\begin{figure}[t]
    \begin{minipage}[t]{0.48\textwidth}
        \hfill 
        \includegraphics[width=\linewidth]{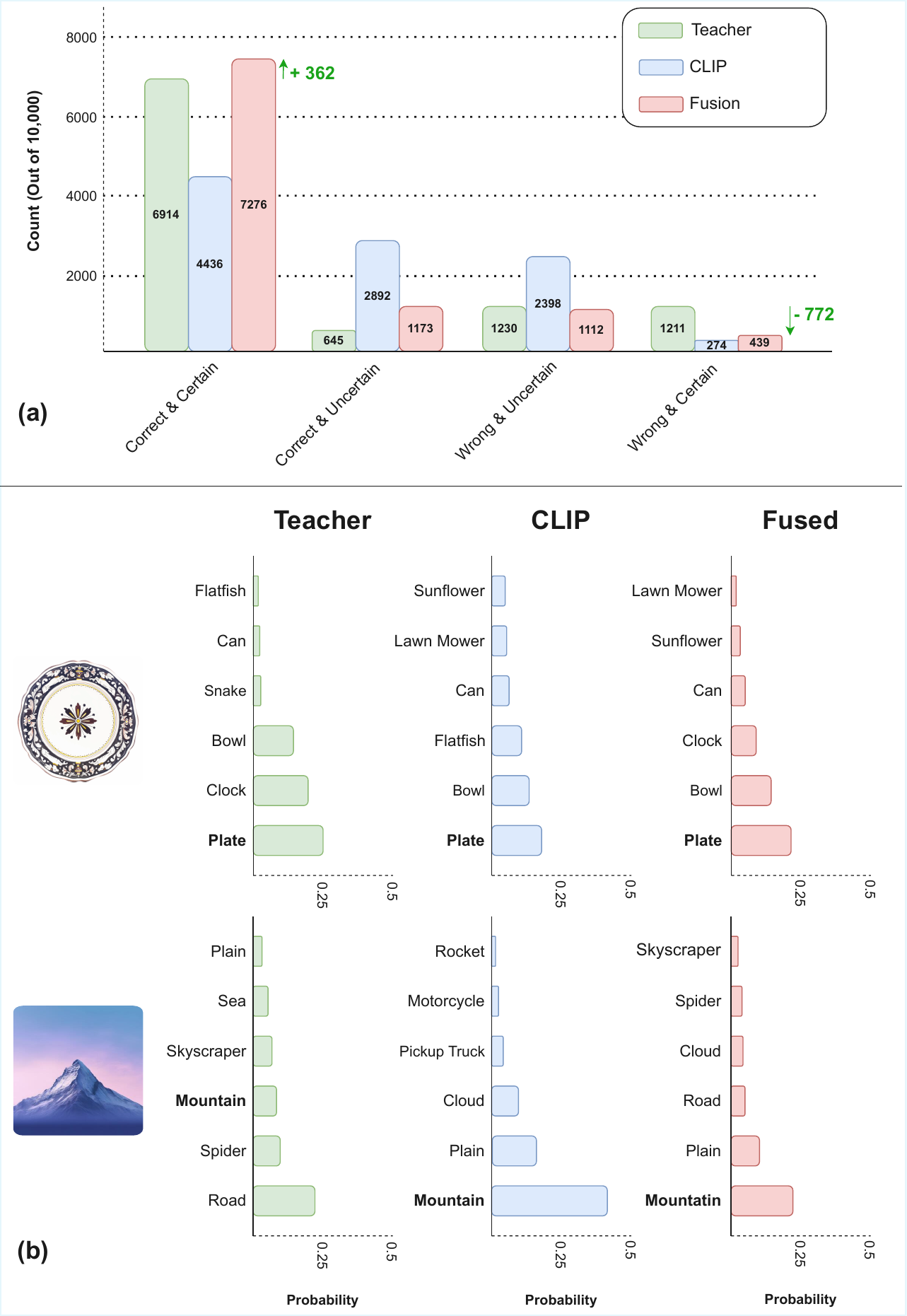}
        \caption{\textbf{Impact of cross-modal teacher fusion on CIFAR-100.} (a) Effect of perturbing the logits of the conventional teacher with CLIP’s logits across four categories, 
considering cases where the teacher’s predictions are correct/incorrect and certain/uncertain. (b) Effect of fusion with CLIP for two sample cases: when the teacher is incorrect, and when the teacher is correct but uncertain.
}
        \label{fig:pull_figure}
    \end{minipage}
\end{figure}

In contrast, large-scale vision–language models such as CLIP~\cite{radford2021learning} capture rich cross-modal knowledge by jointly learning from hundreds of millions of image–text pairs. CLIP’s training paradigm grounds visual concepts in linguistic semantics, enabling zero-shot classification and robust generalization across diverse datasets. However, while CLIP has revolutionized transfer and retrieval tasks, its potential as a \emph{cross-modal teacher} for knowledge distillation remains largely underexplored~\cite{zhou2025improving}. Integrating CLIP’s multimodal understanding into a KD framework presents a promising avenue for transferring both dataset-specific and semantically enriched knowledge to compact visual students.

This paper introduces En\textbf{Rich}ing \textbf{K}nowledge \textbf{D}istillation (RichKD), a novel cross-modal distillation framework that fuses the loits and features of a conventional teacher with those of a pre-trained CLIP model. Our key insight is that CLIP’s predictions, \textbf{though not tailored to the target dataset}, encapsulate complementary information in their broader semantic space. By fusing CLIP’s logits and features with those of the primary teacher, we effectively inject meaningful cross-modal perturbations into the supervision signal. This fusion acts as a soft ensemble of two diverse teachers; one grounded in dataset-specific discriminative patterns, and the other guided by semantic relationships learned from large-scale vision–language data. Figure \ref{fig:pull_figure}(a) illustrates the number of certain/uncertain and correct/incorrect predictions for the conventional teacher, CLIP, and their fusion. The fusion improves overall accuracy, increasing confident correct predictions by 3.6\% and reducing confident mistakes by 7.7\%. Figure \ref{fig:pull_figure}(b) presents two examples illustrating that perturbing the teacher’s logits with CLIP’s helps correct wrong predictions and refine the confidence distribution by better ranking non-target classes.

To enrich the multimodal supervision, we further employ a \emph{multi-prompting strategy}, generating multiple textual templates per class through CLIP’s text encoder and averaging the corresponding predictions. This multi-prompt ensemble captures different linguistic perspectives, smoothing the supervision and mitigating overfitting to specific textual formulations. The fused signals are then distilled into the student through a logit distillation loss.


In summary, this paper makes the following contributions:
\begin{itemize}
    \item We investigate the integration of cross-modal knowledge into conventional knowledge distillation, an underexplored direction that leverages CLIP’s vision–language representations to enrich the teacher–student learning process.

    
    \item We propose a simple yet effective \emph{logit–feature fusion} mechanism that combines a dataset-specific teacher with CLIP’s multi-prompt predictions to construct semantically diverse and informative supervisory signals.


    \item We provide both theoretical and empirical evidence that this cross-modal fusion enhances the diversity of teacher guidance, yielding improved performance and robustness compared to unimodal KD methods.

    
\end{itemize}


\section{Related Work}
\label{sec:related}
In this section, literature related to this work, including KD and multi-teacher KD, is presented.
\subsection{Knowledge Distillation}
Knowledge distillation was popularized by KD~\cite{hinton2015distilling}, who showed that a compact student can learn from a larger teacher by matching softened teacher logits. Following KD, FitNet~\cite{romero2014fitnets} proposed to distill intermediate features, while RKD~\cite{park2019relational} focused on transferring relational knowledge to the student. Subsequent works can be categorized into three groups: logit-based~\cite{guo2020reducing, zhao2022decoupled, hao2023revisit, miles2024understanding, wei2024scaled, cui2024decoupled, sarridis2025indistill, li2025adaptive, wang2025debiased}, feature-based~\cite{zagoruyko2016paying, heo2019comprehensive, chen2021distilling, guo2023class, mansourian2024attention, liu2024rethinking}, and similarity-based~\cite{tung2019similarity, wang2020intra, liu2021exploring, yang2022cross, huang2022knowledge, xin2024new} distillation methods. Furthermore, some studies have sought to enhance the distillation process using various techniques, such as adaptive distillation~\cite{zhou2020channel, park2020knowledge, chen2021cross, park2022prune, huang2025class, mansourian2025aicsd} or modifying the teacher’s logits using label information~\cite{wen2021preparing, cao2023excellent, lan2025improve, sun2025knowledge}.

Although feature-based methods such as CRD~\cite{tian2019contrastive} and SimKD~\cite{chen2022knowledge} have achieved strong performance in KD, the dominant paradigm for classification remains logit distillation. Numerous logit-based methods have been proposed to improve the vanilla KD approach~\cite{zhao2022decoupled, jin2023multi, li2023curriculum, chi2023normkd, sun2024logit}. For example, DKD~\cite{zhao2022decoupled} decouples the KD loss into target and non-target class components; MLKD~\cite{jin2023multi} introduces multi-level logit distillation; and CTKD~\cite{li2023curriculum} adopts a curriculum temperature strategy for logit distillation. More recently, NormKD~\cite{chi2023normkd} customizes the temperature for each sample, and LSKD~\cite{sun2024logit} applies Z-score standardization to logits before the softmax operation. 

\begin{figure*}[ht]
	\centerline{\includegraphics[width=0.95\textwidth]{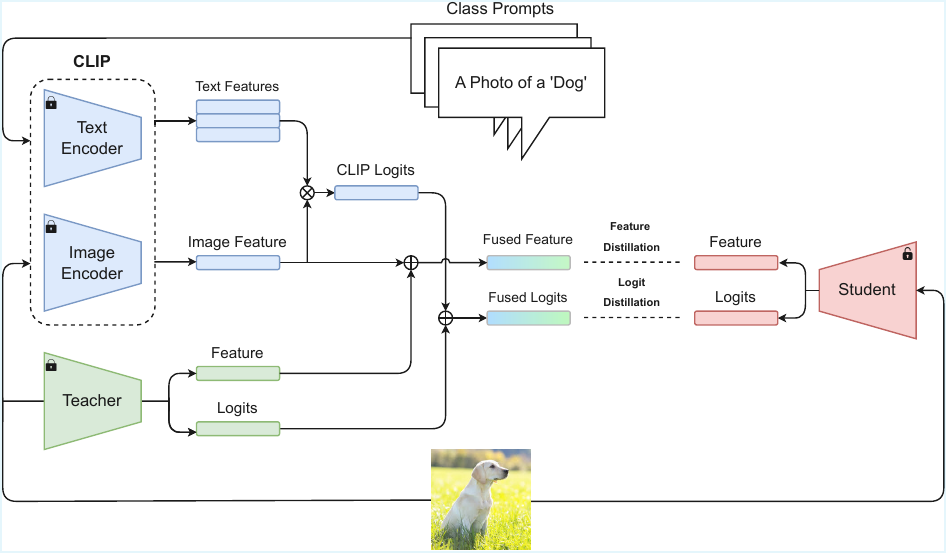}}
	\caption{\textbf{Overall diagram of the proposed RichKD distillation method}. CLIP’s logits and features are fused with those from the conventional teacher model. Feature and logit distillation losses are then defined between the fused representations and the student’s corresponding features and logits. During training phase, the parameters of CLIP and the teacher model are frozen, and the student is trained using feature and logit distillation losses in addition to the cross-entropy loss. Inconsistencies in feature dimensions are addressed through a linear layer transformation.}
	\label{fig:method}
\end{figure*}

\subsection{Multi-Teacher Distillation}
Using multiple teachers or auxiliary teacher assistants is a natural approach to provide richer supervision and to reduce the teacher–student capacity gap. Several strategies have been employed for this purpose, such as averaging logits~\cite{hinton2015distilling}, using voting strategies~\cite{you2017learning, zhang2023adaptive}, perturbing logits~\cite{sau2016deep, hossain2025single}, or assigning teachers that specialize in different subsets of the dataset~\cite{iordache2025multi}. TAKD~\cite{mirzadeh2020improved} was one of the early works that introduced the concept of a teacher assistant, while CA-MKD~\cite{zhang2022confidence} proposed a confidence-aware multi-teacher distillation framework, and DGKD~\cite{son2021densely} presented a densely guided multi-teacher approach. More recently, MLDF~\cite{iordache2025multi} proposed multi-level feature distillation, in which each teacher is specialized on a distinct dataset, and TeKAP~\cite{hossain2025single} introduced an augmentation-based technique that generates multiple synthetic teacher representations by perturbing the knowledge of a single pretrained teacher.

Despite the advancements in multi-teacher distillation, existing methods still lack knowledge diversity, as they rely solely on unimodal visual information. Although works such as CLIP-KD~\cite{yang2024clip} have explored distilling CLIP itself into smaller CLIP model, the use of vision–language models like CLIP as auxiliary teachers in traditional visual knowledge distillation remains largely underexplored. CIKD~\cite{zhou2025improving} was the first to investigate the use of CLIP for knowledge distillation, combining the text features from CLIP with the intermediate layer features of the student to generate new logit outputs.

Our work builds upon these lines of research by introducing a method to inject cross-modal diversity into the distillation process. Instead of training multiple task-specific teachers~\cite{iordache2025multi} or relying solely on perturbation-based approaches~\cite{hossain2025single}, we fuse the logits/features of conventional teacher with a frozen CLIP model to construct an implicit multi-teacher approach that integrates both dataset-specialized and language-grounded knowledge. Furthermore, unlike CIKD~\cite{zhou2025improving}, which introduces additional loss terms based on CLIP’s logits, the proposed method jointly utilizes both the logits and features of CLIP by fusing them with those of the original teacher.

\section{Proposed Method}
\label{sec:method}

In this section, the proposed \textbf{RichKD} is presented, a framework that unifies a task-specific teacher $T$ and the general-purpose CLIP teacher $C$ into a single distillation process. RichKD leverages the complementary strengths of $T$’s dataset-specific discrimination and $C$’s broad semantic knowledge by fusing their logits and features through weighted combinations. Figure \ref{fig:method} shows the overall diagram of the proposed method. This fusion, inspired by ensemble learning theory, reduces individual biases and enhances the generalization of the distilled student.

\subsection{Multi-Prompt CLIP Logits}
Given an input image $x$, the conventional teacher produces a logit vector $z_T(x) \in \mathbb{R}^K$ for $K$ classes. The CLIP model consists of an image encoder $C_{\text{img}}(\cdot)$ and a text encoder $C_{\text{text}}(\cdot)$. For each class $c \in \{1, \dots, K\}$ and prompt $p_m$ from a set of $M$ textual templates $\mathcal{P} = \{p_1, \dots, p_M\}$, the text description is formed

\begin{equation}
t_m(c) = p_m(\text{``class }c\text{''}),
\label{eq:tm}
\end{equation}

\noindent and compute the text embedding $h_m(c) = C_{\text{text}}(t_m(c))$. The CLIP logit for class $c$ under prompt $p_m$ is

\begin{equation}
    z_{C}^{(m)}(x)_c
= \tau \cdot
\frac{
\left\langle
C_{\text{img}}(x),\, h_m(c)
\right\rangle
}{
\|C_{\text{img}}(x)\|_2 \,
\|h_m(c)\|_2
},
\end{equation}

\noindent where $\tau$ is a learned temperature scaling factor, \( \langle \cdot , \cdot \rangle \) denotes the standard inner (dot) product, and \( \| \cdot \|_2 \) denotes the L2 norm. The final CLIP logit is the mean over prompts:

\begin{equation}
    z_C(x) = \frac{1}{M} \sum_{m=1}^{M} z_C^{(m)}(x).
\end{equation}

\noindent This averaging introduces diversity via multiple textual contexts, reducing the variance of individual prompt biases and yielding a smoother, more general prediction distribution.

\subsection{Logit Fusion: Perturbed Supervision}
The teacher and CLIP logits are fused as

\begin{equation}
    z_{F}(x) = \alpha \, z_T(x) + (1 - \alpha) \, z_C(x),
    \label{eq:logit_fusion}
\end{equation}
where $\alpha \in [0,1]$ controls the strength of the conventional teacher relative to CLIP. The student logits $z_S(x)$ are then trained to match the fused logits $z_F(x)$ through a softened Kullback–Leibler (KL) divergence loss:
\begin{equation}
    \mathcal{L}_{\text{logit}} = \mathrm{KL}\!\left(
    \sigma\!\left(\frac{z_F(x)}{T_{\text{temp}}}\right)
    \Big\|\,
    \sigma\!\left(\frac{z_S(x)}{T_{\text{temp}}}\right)
    \right),
    \label{eq:kl_loss}
\end{equation}
where $\sigma(\cdot)$ denotes the softmax and $T_{\text{temp}}$ is the temperature for smoothing.

This simple linear combination can be viewed as a stochastic perturbation of the teacher logits. Because $z_C(x)$ comes from a model trained on diverse open-domain data, it behaves as a low-frequency, semantically meaningful perturbation of $z_T(x)$:
\begin{equation}
\begin{aligned}
z_{F}(x) &= z_T(x) + \epsilon(x),\\
\text{where} \quad 
\epsilon(x) &= (1-\alpha)\,[z_C(x) - z_T(x)].
\end{aligned}
\end{equation}

\noindent In general, the two teachers may have different biases with respect to the true target distribution, so $\mathbb{E}[\epsilon(x)]$ need not be zero. However, as long as the teachers are not perfectly aligned, the perturbation $\epsilon(x)$ has non-zero variance and encodes complementary information from CLIP. This variability acts as a semantically meaningful perturbation of $z_T(x)$, providing a richer and more regularized supervision signal for the student.

\subsection{Feature Fusion}
Beyond the output logits, intermediate representations from both teachers are also transferred. Let $f_T(x)$ and $f_C(x)$ denote the feature maps from the last hidden layer of the supervised teacher and CLIP image encoder, respectively. The fused feature is defined as
\begin{equation}
    f_F(x) = \lambda\, f_T(x) + (1 - \lambda)\, f_C(x),
    \label{eq:feature_fusion}
\end{equation}
where $\lambda \in [0,1]$ balances task-specific and general representations. The student feature $f_S(x)$ is trained to approximate $f_F(x)$ through a general feature distillation loss:
\begin{equation}
    \mathcal{L}_{\text{feat}} = \mathcal{D}\!\left(f_S(x), f_F(x)\right),
\end{equation}
where $\mathcal{D}(\cdot,\cdot)$ denotes a generic similarity or alignment loss (e.g., cosine distance, attention transfer, or contrastive loss).

\begin{table*}[t]
\centering
\caption{Top-1 accuracy (\%) of different knowledge distillation methods on the validation set of CIFAR-100. The teacher and student share the same architecture but differ in configuration. We apply our method to both logit-based and feature-based distillation approaches, and use $\Delta$ to indicate the performance gain. Values in \textcolor{blue}{blue} represent minor improvements, while those in \textcolor{red}{red} indicate significant improvements of at least 0.15. The best and second-best results are indicated by \textbf{bold} and \underline{underlining} font, respectively. “L” and “F” denote logit-based and feature-based methods, respectively.}
\label{tab:cifar100-same}
\resizebox{\textwidth}{!}{
\begin{tabular}{l|ccccccc}
\toprule
\multirow{2}{*}{\textbf{Type}} 
& \textbf{ResNet32$\times$4} & \textbf{VGG13} & \textbf{WRN-40-2} & \textbf{WRN-40-2} 
& \textbf{ResNet56} & \textbf{ResNet110} & \textbf{ResNet110} \\
& \textbf{ResNet8$\times$4} & \textbf{VGG8} & \textbf{WRN-40-1} & \textbf{WRN-16-2} 
& \textbf{ResNet20} & \textbf{ResNet32} & \textbf{ResNet20} \\
\midrule
Teacher & 79.42 & 74.64 & 75.61 & 75.61 & 72.34 & 74.31 & 74.31 \\
Student & 72.50 & 70.36 & 71.98 & 73.26 & 69.06 & 71.14 & 69.06 \\

\midrule

FitNet~\cite{romero2014fitnets} & 73.50 & 71.02 & 72.24 & 73.58 & 69.21 & 71.06 & 68.99 \\
AT~\cite{zagoruyko2016paying} & 73.44 & 71.43 & 72.77 & 74.08 & 70.55 & 72.31 & 70.65 \\
RKD~\cite{park2019relational} & 71.90 & 71.48 & 72.22 & 73.35 & 69.61 & 71.82 & 69.25 \\
OFD~\cite{heo2019comprehensive} & 74.95 & 73.95 & 74.33 & 75.24 & 70.98 & 73.23 & 71.29 \\
SimKD~\cite{chen2022knowledge} & \textbf{78.08} & \textbf{74.89} & \underline{74.53} & 75.53 & 71.05 & 73.92 & 71.06 \\
LSKD~\cite{sun2024logit} & 76.62 & 74.36 & 74.37 & \underline{76.11} & \underline{71.43} & \textbf{74.17} & \underline{71.48} \\

\midrule

KD~\cite{hinton2015distilling} & 73.33 & 72.98 & 73.54 & 74.92 & 70.66 & 73.08 & 70.67 \\
+ RichKD (L) & 75.32 & 73.68 & 74.17 & 76.29 & 71.38 & 73.86 & 71.09 \\
$\Delta$ & \textcolor{red}{1.99} & \textcolor{red}{0.70} & \textcolor{red}{0.63} & \textcolor{red}{1.37} & \textcolor{red}{0.72} & \textcolor{red}{0.78} & \textcolor{red}{0.39} \\

\midrule

CRD~\cite{tian2019contrastive} & 75.51 & 73.94 & 74.14 & 75.48 & 71.16 & 73.48 & 71.46 \\
+ RichKD (F) & 76.09 & 74.22 & 74.70 & 75.63 & 71.64 & 73.99 & 71.77 \\
$\Delta$ & \textcolor{red}{0.58} & \textcolor{red}{0.28} & \textcolor{red}{0.56} & \textcolor{blue}{0.15} & \textcolor{red}{0.48} & \textcolor{red}{0.51} & \textcolor{red}{0.31} \\

\midrule

RichKD (L+F) & \underline{76.72} & \underline{74.86} & \textbf{74.73} & \textbf{76.35} & \textbf{72.12} & \underline{74.08} & \textbf{72.11}\\

\bottomrule
\end{tabular}}
\end{table*}

\subsection{Overall Training Objective}
The complete loss combines ground-truth supervision with fused logit and feature distillation:
\begin{equation}
    \mathcal{L} = \mathcal{L}_{\text{CE}}(y, \sigma(z_S)) + 
    \beta\,\mathcal{L}_{\text{logit}} +
    \gamma\,\mathcal{L}_{\text{feat}},
    \label{eq:total_loss}
\end{equation}
where $\beta$ and $\gamma$ weight the respective distillation terms.

\subsection{Theoretical Motivation}
From a theoretical standpoint, fusing $z_T$ and $z_C$ can be interpreted as
constructing an ensemble teacher $\mathcal{T}_F$ whose predictions are a convex
combination of two heterogeneous teachers. This ensemble can be analyzed using a
bias-variance view. Let $\mathrm{Bias}_T$ and $\mathrm{Bias}_C$ denote the
biases of the task-specific teacher and CLIP teacher with respect to the true
target, and let $\mathrm{Var}_T$ and $\mathrm{Var}_C$ be their prediction
variances. The fused teacher has bias

\begin{equation}
\mathrm{Bias}_F = \alpha\,\mathrm{Bias}_T + (1-\alpha)\,\mathrm{Bias}_C.
\label{eq:bias_fusion}
\end{equation}
\noindent and variance

\begin{equation}
\begin{aligned}
\mathrm{Var}[z_F]
&= \alpha^2 \mathrm{Var}_T + (1-\alpha)^2 \mathrm{Var}_C \\
&\quad + 2\alpha(1-\alpha)\,\mathrm{Cov}[z_T,z_C].
\end{aligned}
\label{eq:var_fusion}
\end{equation}

\noindent The corresponding ensemble error can be written as

\begin{equation}
\mathcal{E}_F = \bigl(\mathrm{Bias}_F\bigr)^2 + \mathrm{Var}[z_F] + \sigma^2 ,
\label{eq:ef}
\end{equation}

\noindent where $\sigma^2$ denotes irreducible noise. When the two teachers have
\emph{complementary} biases (e.g., a task-specific teacher that may overfit and a
CLIP teacher that generalizes more broadly) and their prediction errors are not
highly correlated, one can choose $\alpha \in (0,1)$ such that
$\mathcal{E}_F \le \min\{\mathcal{E}_T,\mathcal{E}_C\}$. In the idealized case
of two approximately unbiased teachers with similar variance and weakly
correlated errors, the ensemble error $\mathcal{E}_F$ becomes strictly smaller
than that of each individual teacher. This provides theoretical support for
using the fused teacher $\mathcal{T}_F$ as a more informative and regularized
supervision signal for the student compared to relying on a single teacher.




\section{Experiments}
\label{sec:experiments}
In this section a complete discussion of the experiments, including datasets, baselines, implementation details, and results is provided. In addition, further experiments, ablation studies, and visualizations, such as class imbalance evaluation, Grad-CAM analyses, training time complexity, prompt templates, and experiments with transformer-based architectures, are also presented in the supplementary material.

\noindent \textbf{Datasets.} Experiments are conducted on the \textbf{CIFAR-100}~\cite{krizhevsky2009learning} and \textbf{ImageNet}~\cite{russakovsky2015imagenet} datasets. The \textbf{CIFAR-100} dataset consists of 50{,}000 training and 10{,}000 validation images across 100 classes, with each image having a resolution of 32$\times$32 pixels. The \textbf{ImageNet} dataset is a large-scale benchmark for image classification, containing approximately 1.28 million training and 50{,}000 validation images from 1{,}000 categories. To evaluate the generalization capability of the proposed method, additional experiments are performed on the corrupted versions of \textbf{CIFAR-100} dataset~\cite{hendrycks2018benchmarking},designed to assess model robustness under distributional shifts.

\noindent \textbf{Baselines.}
The effect of the proposed method is evaluated across multiple logit-based and feature -based distillation approaches, 
including \textbf{KD}~\cite{hinton2015distilling}, \textbf{FitNet}~\cite{romero2014fitnets}, \textbf{AT}~\cite{zagoruyko2016paying}, \textbf{RKD}~\cite{park2019relational}, \textbf{OFD}~\cite{heo2019comprehensive}, \textbf{SimKD}~\cite{chen2022knowledge}, \textbf{LSKD}~\cite{sun2024logit}, \textbf{CRD}~\cite{tian2019contrastive}, \textbf{DKD}~\cite{zhao2022decoupled}, \textbf{MLKD}~\cite{jin2023multi}, \textbf{NormKD}~\cite{chi2023normkd}, and \textbf{RLD}~\cite{sun2025knowledge}. 
Comparisons are also made with various multi-teacher KD methods, 
including \textbf{TAKD}~\cite{mirzadeh2020improved}, \textbf{CA-MKD}~\cite{zhang2022confidence}, \textbf{DGKD}~\cite{son2021densely}, \textbf{CIKD}~\cite{zhou2025improving},  and \textbf{TeKAP}~\cite{hossain2025single}.

\noindent \textbf{Implementation Details.} The same experimental settings as previous works~\cite{chen2021distilling, jin2023multi, zhao2022decoupled, sun2024logit} are followed. For experiments on \textbf{CIFAR-100}, the optimizer is set to SGD, and the number of training epochs is 240, except for \textbf{MLKD}, which is trained for 480 epochs. The initial learning rate is set to 0.01 for \textbf{MobileNets} and \textbf{ShuffleNets}, and 0.05 for other architectures, including \textbf{ResNets}, \textbf{WRNs}, and \textbf{VGGs}. All reported results are averaged over four independent trials. More detailed experimental configurations are provided in the supplementary materials.

All hyperparameters were fine-tuned to select optimal values. We set $\alpha$ in Eq.~\eqref{eq:logit_fusion} to $0.7$, $T_{\text{temp}}$ in Eq.~\eqref{eq:kl_loss} to $3$, $\lambda$ in Eq.~\eqref{eq:feature_fusion} to $0.7$, $\beta$ in Eq.~\eqref{eq:total_loss} to $3$, and $\gamma$ in Eq.~\eqref{eq:total_loss} to $0.8$.

\begin{table*}[t]
\centering
  \caption{Top-1 accuracy (\%) of different knowledge distillation methods on the validation set of CIFAR-100. The teacher and student have different architectures. We apply our method to both logit-based and feature-based distillation approaches, and use $\Delta$ to indicate the performance gain. Values in \textcolor{blue}{blue} represent minor improvements, while those in \textcolor{red}{red} indicate significant improvements of at least 0.15. The best and second-best results are indicated by \underline{underlining} and \textbf{bold} font, respectively. “L” and “F” denote logit-based and feature-based methods, respectively.}
  \label{tab:cifar100-distinct}
  \resizebox{\textwidth}{!}{
  \begin{tabular}{l|ccccccc}
    \toprule
    \multirow{2}{*}{\textbf{Type}} 
      & \multicolumn{1}{c}{\textbf{ResNet32$\times$4}} 
      & \multicolumn{1}{c}{\textbf{ResNet32$\times$4}} 
      & \multicolumn{1}{c}{\textbf{ResNet32$\times$4}} 
      & \multicolumn{1}{c}{\textbf{ResNet32$\times$4}}
      & \multicolumn{1}{c}{\textbf{WRN-40-2}}  
      & \multicolumn{1}{c}{\textbf{VGG13}} 
      & \multicolumn{1}{c}{\textbf{ResNet50}} \\
    & \textbf{SHN-V2} & \textbf{SHN-V1} & \textbf{WRN-16-2} & \textbf{WRN-40-2} & \textbf{ResNet8$\times$4} 
      & \textbf{MN-V2} & \textbf{MN-V2} \\
    \midrule
    Teacher & 79.42 & 79.42 & 79.42 & 79.42 & 75.61 & 74.64 & 79.34 \\
    Student & 71.82 & 70.50 & 73.26 & 75.61 & 72.50 & 64.60 & 64.60 \\

\midrule

FitNet~\cite{romero2014fitnets} & 73.54 & 73.59 & 74.70 & 77.69 & 74.61 & 64.16 & 63.16 \\
AT~\cite{zagoruyko2016paying} & 72.73 & 71.73 & 73.91 & 77.43 & 74.11 & 59.40 & 58.58 \\
RKD~\cite{park2019relational} & 73.21 & 72.28 & 74.86 & 77.82 & 75.26 & 64.52 & 64.43 \\
OFD~\cite{heo2019comprehensive} & 76.82 & 75.96 & 76.17 & \underline{79.25} & 74.36 & \underline{69.48} & 69.04 \\
SimKD~\cite{chen2022knowledge} & \textbf{78.39} & \textbf{76.31} & \textbf{77.17} & \textbf{79.29} & 75.29 & 69.44 & \underline{69.97} \\
LSKD~\cite{sun2024logit} & 75.56 & - & 75.26 & 77.92 & \textbf{77.11} & 68.61 & 69.02
 \\

\midrule

KD~\cite{hinton2015distilling} & 74.45 & 74.07 & 74.90 & 77.70 & 73.97 & 67.37 & 67.35 \\
+ RichKD (L) & 75.68 & 75.04 & 75.39 & 77.82 & 75.77 & 69.03 & 68.07 \\
$\Delta$ & \textcolor{red}{1.23} & \textcolor{red}{0.97} & \textcolor{red}{0.49} & \textcolor{blue}{0.12} & \textcolor{red}{1.80} & \textcolor{red}{1.66} & \textcolor{red}{0.77} \\

\midrule

CRD~\cite{tian2019contrastive} & 75.65 & 75.11 & 75.65 & 78.15 & 75.24 & 69.73 & 69.11 \\
+ RichKD (F) & 76.49 & 75.65 & 76.53 & 78.67 & 76.28 & 69.87 & 69.81 \\
$\Delta$ & \textcolor{red}{0.84} & \textcolor{red}{0.54} & \textcolor{red}{0.88} & \textcolor{red}{0.52} & \textcolor{red}{1.04} & \textcolor{blue}{0.14} & \textcolor{red}{0.70} \\

\midrule

RichKD (L+F) & \underline{76.95} & \underline{76.13} & \underline{76.58} & 78.76 & \underline{76.31} & \textbf{70.03} & \textbf{70.15}\\

\bottomrule
\end{tabular}}
\end{table*}

\begin{table}[ht]
  \centering
  \caption{Top-1 accuracy (\%) comparison with existing multi-teacher distillation methods on the CIFAR-100 validation set.}
  \label{tab:multi_teacher_comparison}
    \resizebox{\linewidth}{!}{%
  \begin{tabular}{>{\centering\arraybackslash}m{2.5cm}|
                >{\centering\arraybackslash}m{2cm}
                >{\centering\arraybackslash}m{2cm}}
    \toprule
    \multirow{2}{*}{Method} &
    \begin{tabular}[c]{@{}c@{}}\textbf{ResNet32×4} \\ \textbf{ResNet8×4} \end{tabular} &
    \begin{tabular}[c]{@{}c@{}}\textbf{WRN\_40\_2} \\ \textbf{WRN\_40\_1}\end{tabular} \\
    \midrule
    Teacher & 79.42 & 75.61 \\
    Student & 72.50 & 71.98 \\
    \midrule
    TAKD~\cite{mirzadeh2020improved} & 73.93 & 73.83 \\
    CA-MKD~\cite{zhang2022confidence} & 75.90 & 74.56 \\
    DGKD~\cite{son2021densely} & 75.31 & 74.23 \\
    CIKD~\cite{zhou2025improving} & 74.79 & 74.32 \\
    TeKAP~\cite{hossain2025single} & 75.98 & 74.41 \\
    \midrule
    \textbf{RichKD} & \textbf{76.72} & \textbf{74.73} \\
    \bottomrule
  \end{tabular}
  }
\end{table}

\subsection{Main Results}
\textbf{Results on CIFAR-100.} 
The proposed method is compared with several prominent feature- and logit-based approaches using different teacher and student architectures. Table \ref{tab:cifar100-same} shows the results of our method when the teacher and student share a similar architecture, while Table \ref{tab:cifar100-distinct} presents the results for cases where the teacher and student have distinct architectures. It can be seen that RichKD can be combined with KD for logit distillation and CRD for feature distillation, and that their combination consistently achieves comparable or better results to existing methods on different architectures.

As RichKD employs an auxiliary teacher model, it inherently follows a multi-teacher distillation framework. Table \ref{tab:multi_teacher_comparison} presents the results of our method in comparison with existing multi-teacher approaches. It can be seen that our method outperforms popular and recently proposed multi-teacher methods across two different teacher–student architecture settings. It should be noted that although RichKD adopts a multi-teacher framework by incorporating CLIP in addition to the primary teacher, CLIP has not been trained specifically on the target dataset and is merely utilized to inject general knowledge into the teacher.

Furthermore, Table \ref{tab:sota_combination} presents the results of combining our method with recent logit-based distillation approaches, namely DKD, MLKD, NormKD, and RLD, across two different teacher–student architecture settings. It can be seen that RichKD can be seamlessly integrated into each of these methods, further improving their performance.

\textbf{Results on ImageNet.} 
To validate the scalability of the proposed method, the results on the large-scale ImageNet dataset are reported. Table \ref{tab:imagenet_ours_error} presents the results of RichKD in comparison with the KD baseline and the recent TeKAP method, where ResNet-34 and ResNet-18 are used as the teacher and student models, respectively.

\begin{table}[ht]
  \centering
  \caption{Impact of integrating the proposed method with recent logit-based distillation methods in terms of Top-1 accuracy (\%) on the CIFAR-100 validation set.}
  \label{tab:sota_combination}
  \resizebox{\linewidth}{!}{%
  \begin{tabular}{>{\centering\arraybackslash}m{2cm}|
                >{\centering\arraybackslash}m{2.5cm}
                >{\centering\arraybackslash}m{2.5cm}}
    \toprule
    \multirow{2}{*}{Method} &
    \begin{tabular}[c]{@{}c@{}}\textbf{ResNet32×4} \\ \textbf{ResNet8×4} \end{tabular} &
    \begin{tabular}[c]{@{}c@{}}\textbf{WRN\_40\_2} \\ \textbf{WRN\_40\_1}\end{tabular} \\
    \midrule
    Teacher & 79.42 & 75.61 \\
    Student & 72.50 & 71.98 \\
    \midrule
    DKD~\cite{zhao2022decoupled} & 76.32 & 74.81 \\
    + RichKD (L) & 76.78 (+0.46) & 75.49 (+0.68) \\
    \midrule
    MLKD~\cite{jin2023multi} & 77.08 & 75.35 \\
    + RichKD (L) & 77.41 (+0.33) & 75.82 (+0.47) \\
    \midrule
    NormKD~\cite{chi2023normkd} & 76.57 & 74.84 \\
    + RichKD (L) & 76.73 (+0.16) & 74.99 (+0.15) \\
    \midrule
    RLD~\cite{sun2025knowledge} & 76.11 & 74.58 \\
    + RichKD (L) & 76.45 (+0.34) & 74.84 (+0.26) \\
    \bottomrule
  \end{tabular}
  }
\end{table}

\begin{table}[t]
\centering
\caption{Scalability comparison on ImageNet. Lower is better (error \%).}
\label{tab:imagenet_ours_error}
\resizebox{\linewidth}{!}{%
\begin{tabular}{>{\centering\arraybackslash}m{1cm}
                >{\centering\arraybackslash}m{1cm}
                >{\centering\arraybackslash}m{1.2cm}|
                >{\centering\arraybackslash}m{1cm}
                >{\centering\arraybackslash}m{1cm}
                >{\centering\arraybackslash}m{1.8cm}}
\toprule
\textbf{Set} & \textbf{Teacher} & \textbf{Student} & \textbf{KD} & \textbf{TeKAP} & \textbf{RichKD (L)} \\
\midrule
Top-1 & 26.69 & 30.25 & 29.59 & 29.33 & \textbf{29.10} \\
Top-5 &  8.58 & 10.93 & 10.30 & 10.08 &  \textbf{9.85} \\
\bottomrule
\end{tabular}
}
\end{table}

\begin{table}[ht]
  \centering
  \caption{Adversarial robustness of the proposed method. Top-1/Top-5 accuracy are reported for FGSM and PGD attacks with different attack parameters.}
  \label{tab:robustness_sweep_merged}
  \resizebox{\linewidth}{!}{%
  \begin{tabular}{>{\centering\arraybackslash}m{2cm}|
                  >{\centering\arraybackslash}m{2cm}
                  >{\centering\arraybackslash}m{2cm}
                  >{\centering\arraybackslash}m{2cm}}
  \toprule
    \multicolumn{4}{c}{\textbf{FGSM Attack}} \\
    \midrule
    Method & $\epsilon=0.001$ & $\epsilon=0.005$ & $\epsilon=0.010$ \\
    \midrule
    KD & 23.51 / 45.68 & 21.91 / 43.86 & 19.99 / 41.89 \\
    RichKD (L) &  25.75 / 49.21 & 23.68 / 47.44 & 21.70 / 45.30 \\
    \midrule
    \multicolumn{4}{c}{\textbf{PGD Attack}} \\
    \midrule
    Method & $\epsilon=0.001$ & $\epsilon=0.005$ & $\epsilon=0.010$ \\
    \midrule
    KD  & 22.13 / 44.43 & 15.25 / 38.57 & 10.61 / 32.60 \\
    RichKD (L)  & 24.30 / 48.03 & 17.72 / 41.55 & 12.61 / 35.66 \\
    \bottomrule
  \end{tabular}
  }
\end{table}

\begin{table}[t]
\centering
\caption{Robustness comparison (accuracy \%) on corrupted versions of the CIFAR-100 dataset.}
\label{tab:corruptions}
\begin{tabular}{
    l
    S[table-format=2.2] S[table-format=2.2]
    S[table-format=2.2] S[table-format=2.2]
}
\toprule
\multirow{2}{*}{Corruptions} & \multicolumn{2}{c}{\textbf{KD}} & \multicolumn{2}{c}{\textbf{RichKD (L)}} \\
\cmidrule(lr){2-3}\cmidrule(lr){4-5}
& {\textbf{Top-1}} & {\textbf{Top-5}} & {\textbf{Top-1}} & {\textbf{Top-5}} \\
\midrule
gaussian noise &  14.42 & 34.70 & 15.02 & 35.87 \\
motion blur &  48.45 & 49.34 & 49.34 & 75.96 \\
snow &  48.22 & 74.93 & 51.03 & 77.76 \\
jpeg compression &  45.34 & 73.16 & 46.99 & 75.18 \\
spatter &  51.16 & 78.00 & 55.04 & 81.43 \\
\midrule
average &  41.51 & 62.02 & 43.48 & 69.24 \\
\bottomrule
\end{tabular}
\end{table}

\subsection{Generalization Comparison}
To validate the effectiveness of the proposed method, generalization ability of students trained using vanilla KD and RichKD are compared. Table \ref{tab:robustness_sweep_merged} presents the results on adversarially perturbed data generated by two well-known attack methods, FGSM~\cite{goodfellow2014explaining} and PGD~\cite{madry2017towards}, with different noise parameters. Both the teacher and the student are trained on the clean CIFAR-100 dataset, using ResNet-32×4 and ResNet-8×4 architectures, respectively. The results demonstrate that our method achieves higher top-1 and top-5 accuracy, benefiting from the inclusion of CLIP, which provides general knowledge not limited to the target dataset images.

Furthermore, Table \ref{tab:corruptions} shows the reulst of vanilla KD and RichKD on several corrupted versions of the CIFAR-100. All the models are trained on clean CIFAR-100, and teacher and student architecters are ResNet-32×4 and ResNet-8×4, respectively. Similar to Table \ref{tab:robustness_sweep_merged}, it can be observed that the student trained with RichKD is more robust to different types of corruption. The proposed method consistently improves the top-1 and top-5 accuracy of the baseline method and achieves a significant performance margin over the baseline for certain corruption types.

\subsection{Ablation Studies}
To further validate the effectiveness of the proposed method, a series of ablation studies are conducted. Since RichKD employs CLIP as an auxiliary teacher to perturb the logits of the primary teacher, the choice and performance of the CLIP model play an important role in our framework. Table \ref{tab:ablation_CLIPs} presents the results of our method using different variants of CLIP, compared to the vanilla KD method, across three different teacher–student architecture pairs. It can be seen that ViT-L/14 improves the distillation performance more than the other CLIP variants, as it employs a more powerful image encoder and contains a larger number of parameters.

\begin{figure}[t]
    \begin{minipage}[t]{0.48\textwidth}
        \hfill 
        \includegraphics[width=\linewidth]{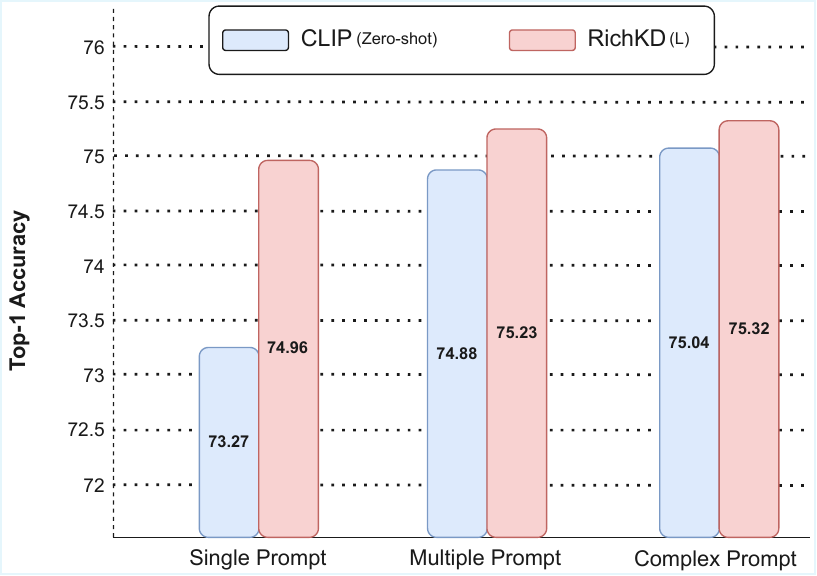}
        \caption{Impact of different types of prompting on CLIP’s zero-shot performance and the student’s performance. The teacher and student architectures are ResNet-32×4 and ResNet-8×4, respectively.}
        \label{fig:prompts}
    \end{minipage}
\end{figure}

Moreover, Figure \ref{fig:prompts} illustrates the effect of different prompting strategies on the distillation results. We evaluate three types of prompts for CLIP’s text encoder: (1) single prompting, where a single template containing the class name is used; (2) multi-prompting, where multiple templates are used with the class name; and (3) complex prompting, where, in addition to multiple prompt templates, several synonyms of each class name are incorporated to further enrich the text embeddings. As shown in the figure, both multi-prompting and complex prompting improve the zero-shot performance of CLIP, which subsequently leads to better distillation results. The detailed prompt formulations are presented in the supplementary material.


\begin{table}[ht]
  \centering
  \caption{Impact of different CLIP variants on distillation performance on the CIFAR‑100 validation set; zero‑shot accuracy for each CLIP model is shown in parentheses.}
  \label{tab:ablation_CLIPs}
  \resizebox{\linewidth}{!}{%
  \begin{tabular}{%
    >{\centering\arraybackslash}m{2.5cm} |  
    >{\centering\arraybackslash}m{2.5cm} |  
    >{\centering\arraybackslash}m{2cm}      
    >{\centering\arraybackslash}m{2cm}      
    >{\centering\arraybackslash}m{2cm}      
  }
    \toprule
    \multirow{2}{*}[1ex]{\textbf{Method}} &
      \multirow{2}{*}[1ex]{\textbf{CLIP Model}} &
      \textbf{ResNet32×4  ResNet8×4} &
      \textbf{WRN‑40‑2  WRN‑16‑2} &
      \textbf{WRN‑40‑2  ResNet32×4} \\
    \midrule
    Teacher & — & 79.42 & 75.61 & 75.61 \\
    Student & — & 72.50 & 73.26 & 72.50 \\
    KD      & — & 73.33 & 74.92 & 73.97 \\
    \midrule
    \multirow{4}{*}{RichKD (L)} &
      RN101 (41.54)   & 74.89 & 75.03 & 75.34 \\
    & RN50x64 (52.15)   & 74.98 & 75.37 & 75.63 \\
    & ViT‑B/32 (61.68)   & 74.75 & 75.79 & 75.49 \\
    & ViT‑L/14 (73.27)   & 75.32 & 76.29 & 75.77 \\
    \bottomrule
  \end{tabular}%
  }
\end{table}

\subsection{Qualitative Results}
\begin{figure}[t]
    \begin{minipage}[t]{0.48\textwidth}
        \hfill 
        \includegraphics[width=\linewidth]{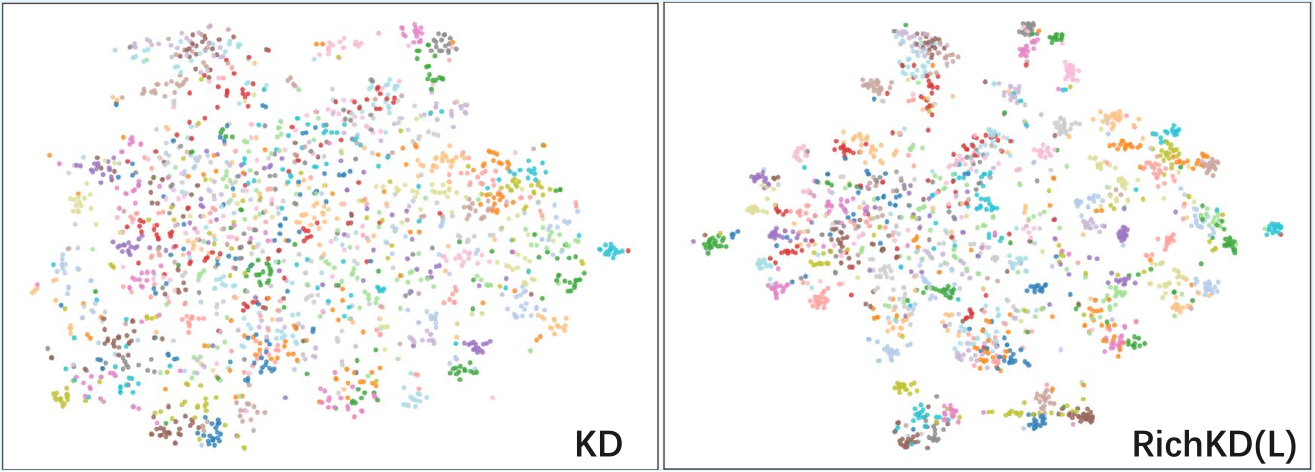}
        \caption{t-SNE visualization of features.}
        \label{fig:tsne}
    \end{minipage}
\end{figure}

In addition to the quantitative comparisons, qualitative results are also presented. Figure \ref{fig:tsne} shows the feature representations projected into a 2D space using t-SNE. The final feature layer of each model is used, taken before the classification head. As can be observed, the embeddings of different classes are more clearly separated in the feature space for the student model trained with RichKD compared to the one trained without CLIP.

\begin{figure}[t]
    \begin{minipage}[t]{0.48\textwidth}
        \hfill 
        \includegraphics[width=\linewidth]{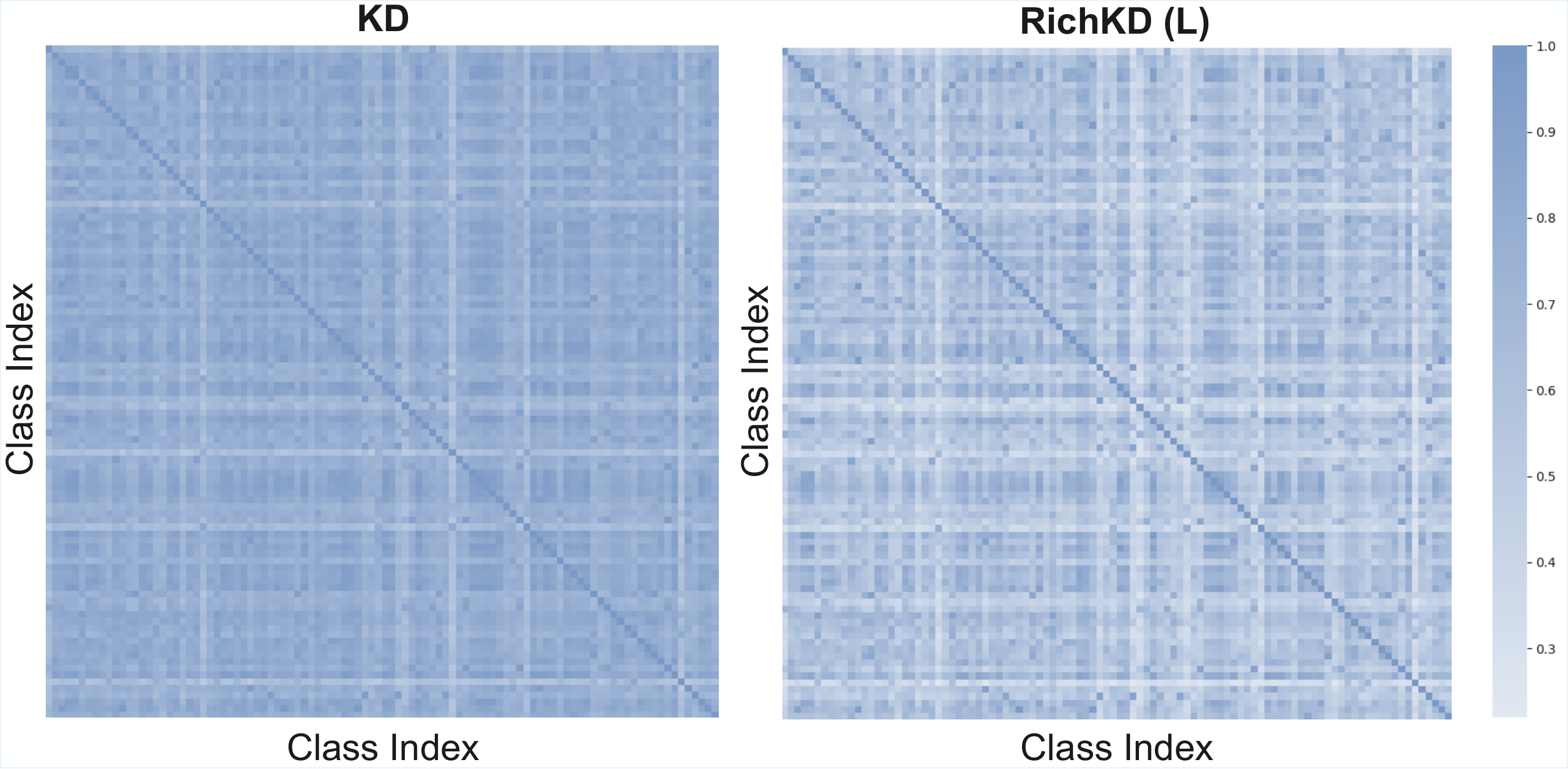}
        \caption{Inter-class correlation matrices on the CIFAR-100 dataset.}
        \label{fig:correlation}
    \end{minipage}
\end{figure}

Furthermore, Figure \ref{fig:correlation} presents a comparison of inter-class correlations between vanilla KD and RichKD. It is evident that inter-class correlations are lower in RichKD, which can be attributed to the influence of CLIP’s logits, which can alter the ranking of the top probabilities, thereby helping to reduce inter-class correlations and improve class separability. For both experiments, ResNet-32×4 and ResNet-8×4 were used as the teacher and student models, respectively. 

\section{Discussion}
\textbf{Limitations.} Using CLIP during training introduces a modest computational overhead due to additional forward passes. Although this occurs only during training, RichKD requires longer training time than standard KD. To reduce this cost, CLIP’s logits and features are cached and reused across epochs, making the extra overhead negligible (see supplementary material for details). Moreover, while RichKD benefits from CLIP’s broad visual–language knowledge, its effectiveness may diminish in domains underrepresented in CLIP’s pretraining (e.g., medical imagery). In such cases, domain-specific CLIP variants can be seamlessly integrated into our framework.



\noindent\textbf{Future Work.} RichKD illustrates the promise of incorporating CLIP’s cross-modal knowledge into knowledge distillation, yet several directions remain open. Our framework currently depends on hand-crafted prompts to query CLIP; integrating advanced techniques such as prompt tuning or dataset-specific textual descriptions could yield stronger supervision. Furthermore, since CLIP provides knowledge complementary to the in-domain teacher, developing adaptive strategies that selectively exploit CLIP’s information per input sample is a promising avenue for future research toward more context-aware distillation.


\section{Conclusion}
In this work, we demonstrated that incorporating CLIP’s cross-modal knowledge as an auxiliary teacher significantly enhanced conventional knowledge distillation. The proposed framework effectively combined dataset-specific and semantically enriched cues, leading to improved student performance, better robustness, and greater generalization. These findings underscored the potential of leveraging vision–language models to enrich traditional visual knowledge distillation approaches.

{
    \small
    \bibliographystyle{ieeenat_fullname}
    \bibliography{main}
}


\end{document}